\documentclass[runningheads]{llncs}
\usepackage[T1]{fontenc}
\usepackage{graphicx}
\usepackage{booktabs}
\usepackage[misc]{ifsym}

\usepackage{mwe}

\usepackage{times}
\usepackage{soul}
\usepackage{url}
\usepackage[hidelinks]{hyperref}
\usepackage[utf8]{inputenc}
\usepackage[small]{caption}
\usepackage{graphicx}
\usepackage{amsmath}
\usepackage{booktabs}
\usepackage{array}
\usepackage{tabularx}
\usepackage{makecell}
\usepackage{diagbox}
\usepackage{multirow}
\usepackage{amsfonts}
\usepackage{latexsym}
\usepackage{adjustbox}
\usepackage{bbold}
\usepackage{wrapfig} 
\usepackage{color}
\usepackage{float}
\usepackage{dblfloatfix}
\begin{document}

\title{Fairness-Aware Graph Representation Learning with Limited Demographic Information}

\titlerunning{\fontsize{8}{10}\selectfont Fairness-Aware Graph Representation Learning with Limited Demographic Information}

\author{Zichong Wang\inst{1} \and 
Zhipeng Yin\inst{1} \and 
Liping Yang\inst{2} \and \\
Jun Zhuang\inst{3} \and
Rui Yu\inst{4} \and 
Qingzhao Kong\inst{1} \and 
Wenbin Zhang \Letter\inst{1}}

\institute{Florida International University, Miami, USA\\ \and
University of New Mexico, Albuquerque, USA\\ \and
Boise State University, Boise, USA\\ \and
University of Louisville, Louisville, USA\\ 
\email{\{ziwang,wenbin.zhang\}@fiu.edu}}

\tocauthor{Zichong Wang, Zhipeng Yin, Liping Yang, Jun Zhuang, Rui Yu, Qingzhao Kong, Wenbin Zhang}

\authorrunning{Zichong Wang et al.}

\maketitle              
 
\begin{abstract}
Ensuring fairness in Graph Neural Networks is fundamental to promoting trustworthy and socially responsible machine learning systems. In response, numerous fair graph learning methods have been proposed in recent years. However, most of them assume full access to demographic information, a requirement rarely met in practice due to privacy, legal, or regulatory restrictions. To this end, this paper introduces a novel fair graph learning framework that mitigates bias in graph learning under limited demographic information. Specifically, we propose a mechanism guided by partial demographic data to generate proxies for demographic information and design a strategy that enforces consistent node embeddings across demographic groups. In addition, we develop an adaptive confidence strategy that dynamically adjusts each node’s contribution to fairness and utility based on prediction confidence. We further provide theoretical analysis demonstrating that our framework, FairGLite, achieves provable upper bounds on group fairness metrics, offering formal guarantees for bias mitigation. Through extensive experiments on multiple datasets and fair graph learning frameworks, we demonstrate the framework's effectiveness in both mitigating bias and maintaining model utility. 


\end{abstract}

\section{Introduction}
\label{sec:introduction}

Graph Neural Networks (GNNs) have become a prevalent approach for handling complex real-world applications, such as healthcare~\cite{an2023comprehensive}, social network analysis~\cite{tabassum2018social}, and recommendation systems~\cite{ko2022survey}. The success of GNNs relies on message-passing mechanisms, which aggregate information from neighboring nodes, effectively capturing both graph structural information and node attribute information~\cite{wang2024advancing,zheng2022graph}. However, despite their successes, GNNs tend to inherit and even exacerbate existing biases from graph data~\cite{wang2023mitigating}, propagating and amplifying unfair patterns embedded in network topology and features. This unintended amplification of societal biases and the potential for discriminatory outcomes have highlighted the urgent need to develop strategies that promote fairness within these systems. To this end, a number of approaches~\cite{wang2025fdgen,zhu2024fair,wang2024individual} have been proposed in recent years, with most relying on complete demographic information to guide fair graph learning.

However, this requirement often does not align with realistic situations, as collecting or explicitly utilizing demographic information (\textit{e.g.}, gender, race) can be restricted or prohibited due to privacy concerns, legal constraints, ethical considerations, or social sensitivity~\cite{krumpal2013determinants,lahoti2020fairness,wang2025towards}. For example, in many real-world graph datasets such as academic collaboration networks or online social platforms, demographic information is often missing, incomplete, or intentionally withheld to protect user privacy, with studies showing that less than 30\% of users voluntarily disclose demographic information~\cite{madden2013teens}. Consequently, fairness-aware graph methods that rely on full demographic information become impractical when only limited data is available. This misalignment between theoretical fairness requirements and real-world constraints significantly limits the applicability and deployment of existing approaches.

To fill this gap, a few works~\cite{yan2020fair,lahoti2020fairness,pelegrina2023statistical} have begun exploring fairness without full demographics. However, most of them focus on i.i.d. data settings and do not account for the relational characteristics of graphs. Therefore, these approaches cannot be readily applied to real-world graph data, leaving fairness-aware graph learning with limited demographic information as an open area of research with unique challenges: \textbf{i) Identifying missing demographic information from limited labels.} In real-world graphs, only a small subset of nodes discloses their demographics. These labels often over-represent favored groups or cluster in specific regions of the graph, making it difficult to train reliable predictors for the missing demographics. Naively using such limited labels risks amplifying bias rather than reducing it. \textbf{ii) Enforcing fairness constraints with uncertain demographic labels.} Enforcing fairness constraints becomes particularly challenging when demographic labels are uncertain or sparse, as such uncertainty can reduce the accuracy of demographic group identification. This complicates the enforcement of fairness constraints, posing additional hurdles in maintaining fairness across different groups while achieving comparable prediction performance. \textbf{iii) Theoretical analysis of fairness guarantees under incomplete demographic information.} Providing rigorous theoretical guarantees for fairness metrics becomes inherently complex when demographic labels are incomplete or uncertain. Without complete demographic information, standard theoretical frameworks for fairness analysis may not directly apply, posing significant challenges in theoretically bounding fairness violations and ensuring reliable fairness guarantees.

\sloppy
 
To bridge this critical gap between theoretical fairness requirements and practical constraints, this paper proposes \textit{Fair Graph Representation Learning with Limited Demographics (FairGLite)} specifically designed to mitigate bias in graph learning algorithms when only limited demographic information is available. \textit{To the best of our knowledge, this is the first framework to address bias in graph learning under limited demographic information while providing theoretical guarantees.} Specifically, FairGLite leverages limited demographic labels to guide a causal-analysis-based encoder that generates proxies for missing demographic information. These proxies are then used to enforce representative constraints, ensuring fairness in node representations while preserving task-relevant information. To further balance fairness and utility, FairGLite employs an adaptive confidence strategy that applies fairness constraints primarily to high-confidence samples, mitigating performance loss while improving fairness on uncertain cases. Finally, we provide a theoretical analysis showing that FairGLite achieves an upper bound on fairness metrics, offering formal guarantees for fair predictions within our framework. The main contributions can be summarized as:

\begin{itemize}
    \item Address the overlooked challenge of fair graph learning with incomplete demographics through a proxy-based framework that enforces fairness with three tailored constraints.
    \item Introduce an adaptive confidence strategy that balances fairness and utility by weighting samples according to classification confidence.
    \item Provide theoretical analysis establishing upper bounds on fairness metrics, offering formal guarantees for fair predictions.
    \item Demonstrate effectiveness on four real-world graph datasets, achieving strong bias mitigation while maintaining utility comparable to state-of-the-art methods.
\end{itemize}

\section{Related Work}
\label{sec:related_work}

\subsection{Fairness with Incomplete Demographic Information}

In recent years, the research community has increasingly focused on developing fair machine learning models that function effectively even when demographic data is incomplete or unavailable~\cite{kenfack2024survey}. Existing approaches primarily fall into two categories: proxy-based methods~\cite{grari2021fairness,pelegrina2023statistical,zhao2022towards} and minimax fairness methods~\cite{chai2022fairness,martinez2021blind,sagawa2019distributionally}; Proxy-based methods attempt to infer missing demographic information from correlated features or partial labels, while minimax fairness methods employ John Rawls' difference principle~\cite{rawls1971theories} to optimize performance for the least advantaged subgroup. However, these techniques are mainly designed for independent and identically distributed (i.i.d.) data, neglecting the complexities introduced by relational or graph-structured data. Consequently, applying these fairness methods directly to graph data remains challenging.

\subsection{Fair Graph Learning}

Extensive efforts have been made to specifically address fairness within graph neural networks by mitigating biases in training data~\cite{ling2023learning,wang2025fg} or by incorporating fairness-aware training frameworks~\cite{zhu2024fair,wang2024advancing}. The core idea behind most of these approaches is the removal of demographics-related information, thereby enforcing GNNs to make decisions independent of the demographic information~\cite{zhang2023censored}. Despite their great success, most existing fair GNNs assume access to predefined demographic information during training, which is impractical in most real-world socially sensitive applications due to privacy, legal, or regulatory restrictions~\cite{ashurst2023fairness}. To this end, initial works have started exploring fair graph learning with incomplete demographics. For instance, FairGNN~\cite{dai2021say} employs a demographic estimator combined with adversarial learning to predict missing demographics and improve fairness. Similarly, FairAC~\cite{guo2023fair} aggregates features from neighbors through an attention mechanism to handle nodes with missing demographics. However, these methods fail to account for uneven demographic disclosure across groups, individuals from advantaged groups may disclose their demographics more readily, while disadvantaged groups may withhold them due to fears of discrimination. Such uneven availability can significantly hinder the effectiveness of these methods.

To address this limitation, FairGLite explicitly incorporates an adaptive confidence strategy designed to selectively impose fairness constraints primarily on nodes with high-confidence demographic predictions, effectively managing uneven demographic availability across different groups. Additionally, FairGLite provides rigorous theoretical analysis to establish fairness guarantees under scenarios with incomplete demographic information. This strategy enhances model fairness while minimizing negative impacts on overall model utility.

\section{Notations}
\label{sec:notation}

For clarity, the method and proofs are presented under a node classification setting with binary demographic information and binary labels. Specifically, a graph is represented as $\mathcal{G} = (\mathcal{V}, \mathcal{E}, \mathbf{X})$, where $\lvert \mathcal{V} \rvert = n$ is the number of nodes and $\lvert \mathcal{E} \rvert = r$ is the number of edges. The matrix $\mathbf{X} \in \mathbb{R}^{n \times d}$ contains $d$-dimensional feature vectors, with the $u^{th}$ row corresponding to node $u$. The adjacency matrix $\mathbf{A} \in \{0, 1\}^{n \times n}$ has entries $\mathbf{A}_{u,k} = 1$ if there is an edge $e_{u,k} \in \mathcal{E}$ between nodes $u$ and $k$, and $\mathbf{A}_{u,k} = 0$ otherwise. In addition, $S \in \{0, 1\}^{n \times 1}$ denotes the demographic information while $s_u$ for the value of $u$. Furthermore, $S_{d} = \{\,u \mid s_u = 0\}$ represents the deprived group (for example, female), and $S_{f} = \{\,u \mid s_u = 1\}$ refers to the favored group (for example, male). Each node $u$ also has a one-hot ground-truth label $y_u$, and $\hat{y}_u$ is its predicted label. Lastly, $y_u = 1$ indicates a granted label and $y_u = 0$ denotes a rejected label.

\nocite{wang2023mitigating,wang2024advancing,wang2025fair,wang2025fdgen,zhang2025fairness,zhang2024ai,wang2025towards, wang2023fg2an, wang2023mitigating, chinta2023optimization, chu2024fairness, wang2024advancing, wang2024towards1, yazdani2025generative, amon2024uncertain, yin2025Uncertain, wang2025FairnessT, yin2024improving, chinta2024fairaied, doan2024fairness1, yin2024accessible, wang2025fg, wang2025graph, saxena2023missed, yin2025digital, chinta2025ai, wang2025fdgen, wang2025Redefining, zhang2020flexible, zhang2021fair, zhang2022fairness, zhang2021farf, zhang2023fairness, zhang2016using, zhang2018content, zhang2018deterministic, tang2021interpretable, zhang2021autoencoder, zhang2020learning, zhang2021disentangled, liu2021research, cai2023exploring, guyet2022incremental, zhang2024fairness, liu2023segdroid, zhang2020online, zhang2019faht, zhang2025online, zhang2024ai, zhang2022longitudinal, zhang2025fairness, wang2025towards2, yinAMCR2025, Wang2025Unified, ijcai2025p63, ijcai2025p64, zhang2025datasets, palikhe2025towards, wang2025ai,  wang2024individual1, wang2024individual, wang2025fair, zhang2023censored, wang2026guic,wang2026fair}

\vspace{-0.35cm}
\begin{figure*}[h]
  \centering
  \includegraphics[width=0.6\textwidth]{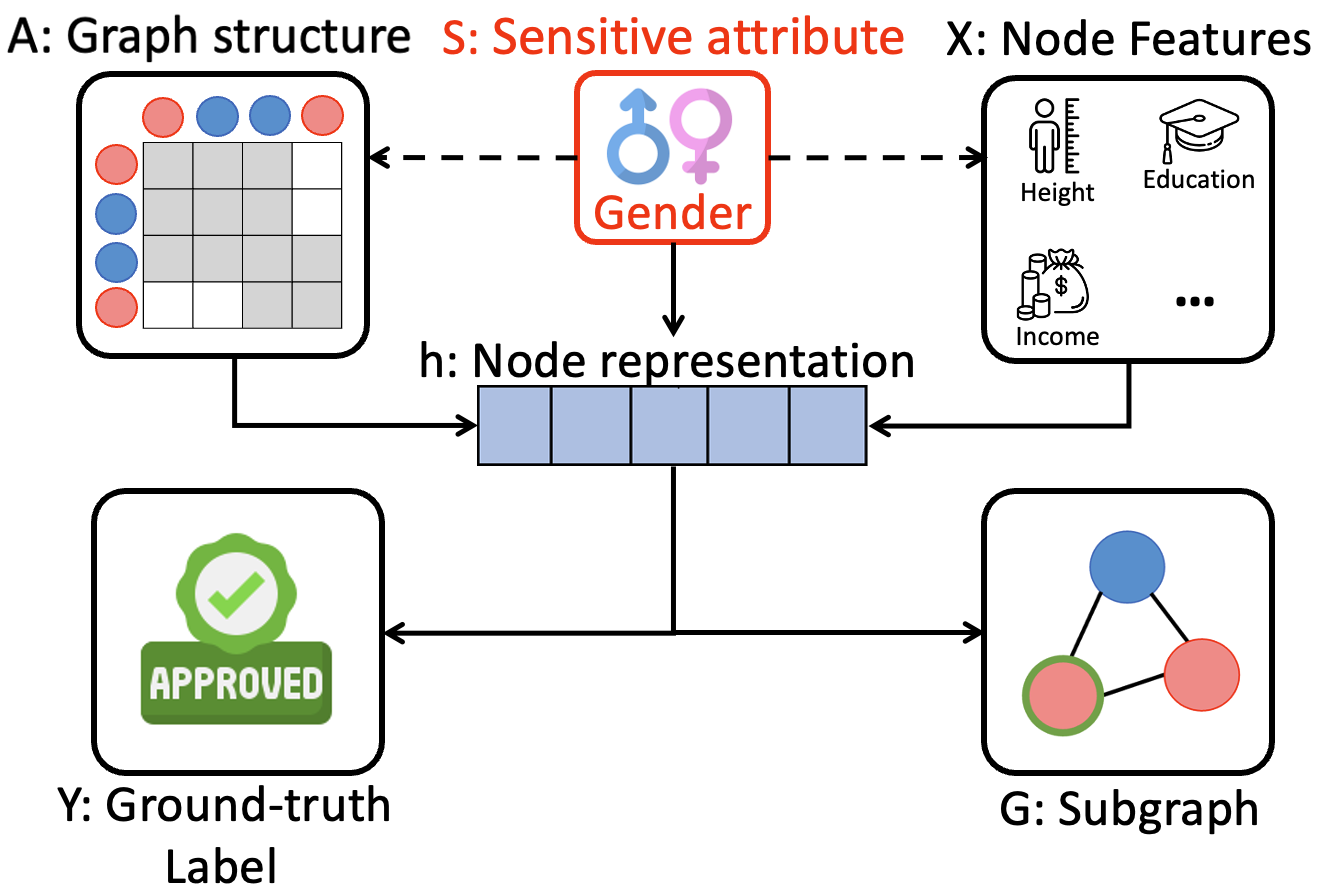}
  \caption{Structural Causal Model for FairGLite.}
  \label{fig:causal}
\end{figure*}
\vspace{-0.9cm}

\section{The Proposed Framework - FairGLite}
\label{sec:method}

\subsection{Fair Causal Analysis}
\label{sec:Fair_Causal_Analysis}

Existing works often rely on correlated non-demographic attributes as proxies for missing demographic information based on prior knowledge~\cite{kenfack2024survey}. However, identifying accurate proxies from high-dimensional non-identically distributed data is both challenging and critical for ensuring fairness. To address this, we employ causal analysis to characterize the underlying mechanisms in the observed graph explicitly. We focus on fair node classification without full demographic information and construct a Structural Causal Model (SCM)~\cite{pearl2000models}, as shown in Figure~\ref{fig:causal}. The SCM captures the causal relationships among six variables: Demographic Information ($S$), Ground-Truth Label ($Y$), Graph Structure ($A$), Node Features ($X$), Ego-graph ($G$), and Node Representation ($h$). Each edge in the SCM represents a causal link. Specifically, $S$ is determined at birth and thus has no parent variable, but influences other variables, including $X$ and $A$. Furthermore, $S$, $X$, and $A$ collectively affect the final node representation through the GNN message-passing mechanism, which should also preserve task-relevant information for downstream node classification and ego-graph reconstruction.

\subsection{FairGLite Framework}
\label{sec:overview}

\begin{figure*}[h]
	\centering
 	\includegraphics[width=1.0\textwidth]{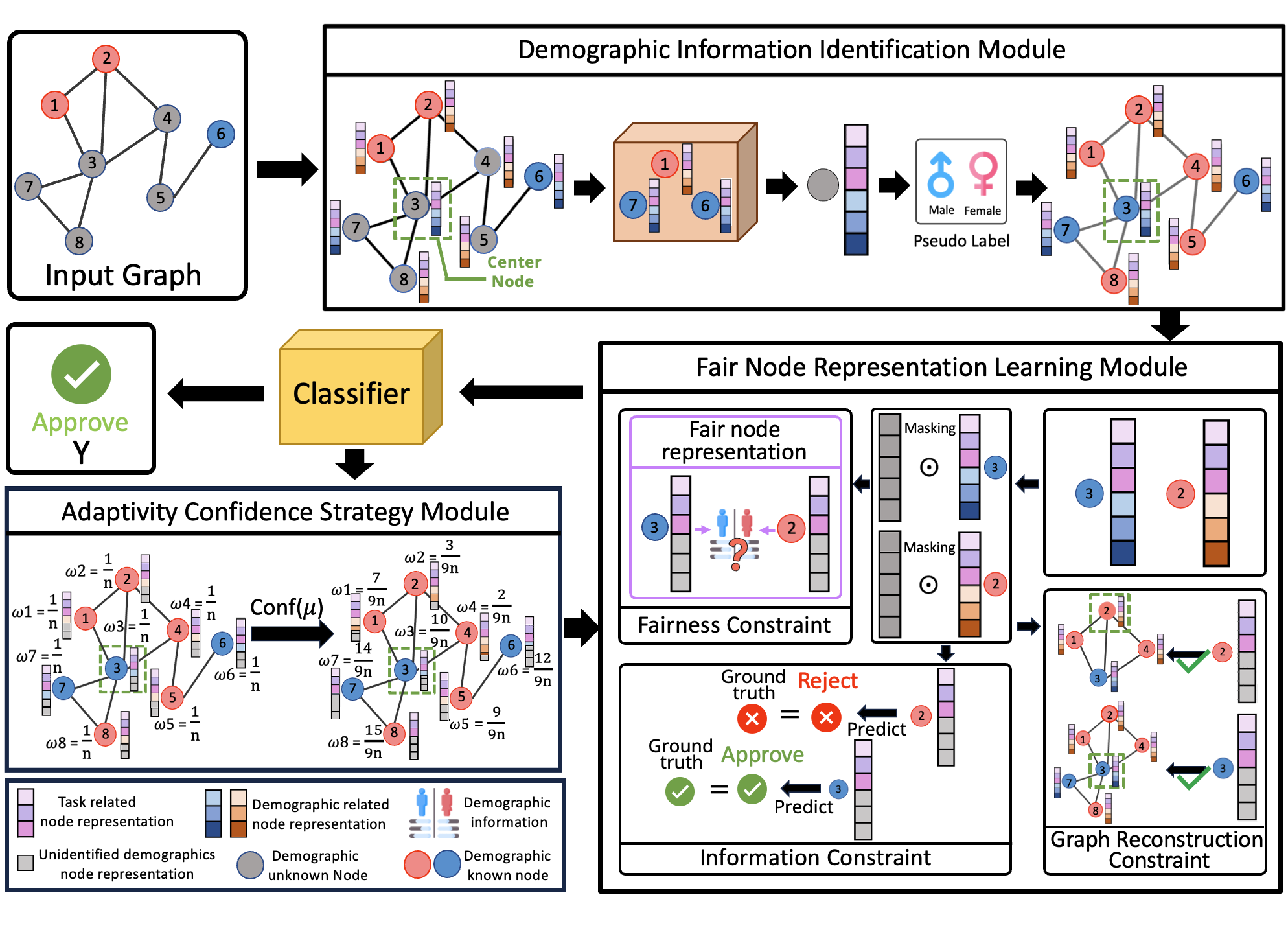}
	\caption{An illustration of the proposed framework FairGLite.}
	\label{fig:overview}
\end{figure*}

Guided by the proposed fair causal model, FairGLite continues to learn fair node representations despite limited access to demographic information. As illustrated in Figure~\ref{fig:overview}, FairGLite operates through three interconnected modules: (i) a demographic information identification module, (ii) a fair node representation learning module, and (iii) an adaptivity confidence strategy module. Initially, the input graph, where nodes are marked with known or unknown demographic labels, is passed to the demographic information identification module. Here, node representations are generated by aggregating information from neighboring nodes' structural features and non-sensitive attributes, forming demographic proxies. These proxies assign pseudo demographic labels to nodes with initially unknown labels, as visually indicated in Figure~\ref{fig:overview}. Next, these demographic proxies are utilized in the fair node representation learning module. FairGLite generates node embeddings subject to three constraints, visually represented by different stages in the figure: the fairness constraint applies learnable masking vectors, depicted as bars beside the nodes, to reduce demographic identifiability in embeddings; the information constraint preserves predictive information, and the graph reconstruction constraint ensures the ego-graph structural integrity of the node embeddings, illustrated by maintaining graph connections. Finally, these node embeddings are passed into the adaptivity confidence strategy module, where nodes are weighted based on the confidence of demographic identification, clearly represented by adjusted node weights in Figure~\ref{fig:overview}. Higher-confidence nodes receive stronger fairness enforcement, depicted by changes in node weight values in Figure~\ref{fig:overview}, thus selectively emphasizing fairness constraints. Additionally, FairGLite is theoretically supported by deriving upper bounds on fairness metrics, providing formal guarantees that reducing representation disparities leads to improved fairness outcomes in downstream tasks. Each of these modules will be detailed in the following discussion.

\subsection{Demographic Information Identification Module}
\label{sec:Demographic_Information_Identification_Module}

This module aims to infer missing demographic labels using the observed graph data (\textit{i.e.}, $X$ and $A$) in conjunction with partially available demographic labels. As demographic information is assumed to influence both graph structure and node features, and to model this effect, we construct a proxy by integrating representations of the graph structure with the node features. Specifically, a graph encoder is employed to generate this proxy, as it is capable of capturing complex patterns in high-dimensional non-identically distributed data and combining information from graph structure and features, even when only a limited number of demographic labels are available. Formally, the encoder is defined as follows:

\begin{equation}
\label{eq:gnn_encoder}
\mathbf{h}_{u}^{(l)} 
\;=\; 
\xi\,\mathbf{h}_{u}^{(l-1)} 
\;+\; 
\sum_{k \in \mathcal{N}(u)} 
\alpha_{u,k}^{(l)} 
\,\mathrm{ReLU}\!\Bigl(\mathbf{W}^{(l)} \,\mathbf{h}_{k}^{(l-1)}\Bigr), \quad
\alpha_{u,k}^{(l)} 
\;=\; 
\frac{\exp \bigl(e_{u,k}^{(l)}\bigr)}
{\sum_{k \in \mathcal{N}(u)} \exp \bigl(e_{u,k}^{(l)}\bigr)}
\end{equation}

\noindent where $\mathbf{h}_{u}^{(l)}$ represents the embedding of node $u$ at layer $l$, and $\mathbf{h}_{u}^{(l-1)}$ is the embedding from the previous layer. In addition, the parameter $\xi$ is a learnable scalar that controls how much of the previous representation is retained at layer $l$, the matrix $\mathbf{W}^{(l)}$ is a learnable weight matrix, and $\mathrm{ReLU}(\cdot)$ is the nonlinear activation function. The set $\mathcal{N}(u)$ denotes the neighborhood of the center node $u$, representing all nodes directly connected to $u$ in the graph. The attention coefficient $\alpha_{u,k}^{(l)}$ reflects the relative importance of neighbor $k$ to node $u$ in the aggregation process.

Based on the encoder architecture, the module is trained via a supervised classification task aimed at predicting missing demographic information. The encoder parameters are learned by minimizing the cross-entropy loss between predicted and observed demographic labels. This supervision encourages the encoder to map nodes with similar demographics to nearby points in the representation space, even when partial label availability is present. Once training converges, the encoder parameters are fixed and subsequently used as a feature extractor that transforms high-dimensional node features into compact, informative embeddings. These low-dimensional representations function as proxies for demographic information in later components of the framework. This approach enables the model to indirectly capture and regulate the effects of missing demographic attributes, despite the absence of complete ground-truth labels.

\subsection{Fair Node Representation Learning Module}
\label{sec:Fair_Node_Representation_Learning_Module}

With demographic proxies obtained from the identification module, the fair node representation learning module tackles the challenge of producing embeddings that remain predictive for downstream tasks while mitigating bias. As discussed earlier, nodes sharing the same demographic information often tend to be more densely connected, creating homophilic clusters within the graph. During message-passing, this structural bias becomes encoded into node representations, as the aggregation process smooths node representations among nodes sharing the same demographic information while amplifying differences across groups. Hence, the learned node representations become overly associated with demographic information. To address this issue, we propose a fair learning strategy based on learnable feature masking. Our approach generates fair node representations by applying a learnable masking vector to original node features, effectively filtering out demographic-related information while preserving task-relevant information.  Specifically, we utilize a learnable masking mechanism to minimize disparities in node representations between subpopulations with different demographic information. This masking approach aims to produce fair node representations that satisfy three key objectives: i) Demographic information becomes less identifiable from the transformed representations, ii) Task-relevant information is preserved for accurate downstream predictions, and iii) Graph structural relationships are maintained to ensure representation quality. To achieve these objectives simultaneously, we design a multi-constraint optimization framework with three complementary components:

\noindent \textbf{Fairness Constraint.} The fairness constraint aims to minimize representation disparities between different demographic subgroups. To formalize this objective, we first establish a theoretical analysis for measuring bias in node representations, then develop a masking-based approach to mitigate such disparities. Specifically, we quantify representation bias as the disparity between group-averaged embeddings as follows:

\begin{equation}
    \Delta_{\rm{bias}} = \left\|\frac{1}{|S_d|} \sum_{v_i \in S_d} \mathbf{h}_i - \frac{1}{|S_f|} \sum_{v_j \in S_f} \mathbf{h}_j\right\|_2
\end{equation}

To further measure this bias in the context of GNN message-passing, we introduce a representation constraint on node embeddings. Specifically, each node $v_i$ have a node representation $h_i^{(l)} = [z_1, z_2, \ldots, z_{d_h}]$ subject to $\mu^{(d)} - \Delta^l \preceq h_i^{(l)} \preceq \mu^{(d)} + \Delta^l$~\cite{kose2024fairgat}, where the parameter $\Delta^l$ serves as a tolerance per layer indicating the allowed deviation of the representation from group mean ($\mu^{(d)}$) along each coordinate. Hence, we quantify the node representation disparity between demographic subgroups during GNN message-passing, which can be upper-bounded as follows:

\begin{align}
\label{equ:biasbound}
    \mathbf{h}_D^{(l)} &= \left\| \frac{1}{N_d} \sum_{i \in \mathcal{S}_d} \mathbf{h}_i^{(l)} - \frac{1}{N_f} \sum_{j \in \mathcal{S}_f} \mathbf{h}_j^{(l)} \right\| \\ \nonumber
    &\leq \left( 3 - \frac{1}{N_d N_f} \sum_{i \in \mathcal{S}_d} \sum_{u \in \mathcal{S}_f} k\left(\sigma(\mathbf{W}^{(l)} \mathbf{h}_u^{(l-1)}), \sigma(\mathbf{W}^{(l)} \mathbf{h}_i^{(l-1)})\right) \right. \nonumber \\ \nonumber
&\quad \left. - \frac{1}{N_f N_d} \sum_{j \in \mathcal{S}_f} \sum_{u \in \mathcal{S}_d} k\left(\sigma(\mathbf{W}^{(l)} \mathbf{h}_u^{(l-1)}), \sigma(\mathbf{W}^{(l)} \mathbf{h}_j^{(l-1)})\right) \right) \left\| \mu_{l-1}^{(d)} - \mu_{l-1}^{(f)} \right\| \nonumber \\ \nonumber
&\quad + \left\| \mu^{(d)} - \mu^{(f)} \right\|  + \left[ C^{(l)} \cdot \frac{1}{4} \left\|W^{(l)}\right\|_2 \left( \left(1 + \frac{2}{N_f}\right) \sqrt{d_h} \Delta^{l-1} + \Delta_{\text{base}}^{(l)} \right) \right]
\end{align}

\noindent where the function $\sigma(\cdot)$ is the activation (\textit{e.g.}, sigmoid), and the constant $C^{(K)} \geq 0$ absorbs fixed numerical factors. In addition, $\Delta^{l-1}$ denotes the per-coordinate tolerance at layer $l - 1$, and $\Delta_{\text{base}}^{(l)}$ collects layer-$l$ approximation slack. What's more, the $N_d$ and $N_f$ are the number of nodes belonging to the favored and deprived groups.

\setcounter{footnote}{0}
Building on the node representation disparity shown in Equation~\ref{equ:biasbound}, we establish an upper bound on the demographic parity violation, providing theoretical guarantees that connect the bias measure to downstream fairness metrics, as detailed in Theorem 4.1 (proof in appendix\footnote{\label{fn:link}\url{https://zichongwang.com/files/FairGLiteAppendix.pdf}}):

\medskip \noindent \textbf{Theorem 4.1} In a node classification task, minimizing node representation discrepancies between two demographic subgroups bounds the disparity as follows:

\begin{equation} 
\scalebox{0.90}{$ 
\begin{aligned} 
\label{eq:DP-GAT-combined}
\Delta_{\text{DP}}
&\;=\;
\Bigl|
\tfrac{1}{|V_{S_d}|}\!\sum_{i\in \mathcal{S}_d} f\bigl(\mathbf{z}_i\bigr)_1
-
\tfrac{1}{|V_{S_f}|}\!\sum_{j\in \mathcal{S}_f} f\bigl(\mathbf{z}_j\bigr)_1
\Bigr| \\
&\;\le\;
 \left|f(\mathbf{z}_{\mu^{(d)}})_1 - f(\mathbf{z}_{\mu^{(f)}})_1\right| + \frac{L}{2} \left( \frac{1}{N_d} \sum_i \left\|\mathbf{z}_i - \mathbf{z}_{\mu^{(d)}}\right\| + \frac{1}{N_f} \sum_j \left\|\mathbf{z}_j - \mathbf{z}_{\mu^{(f)}}\right\| \right)
\end{aligned} $}
\end{equation}

\noindent where $\mathbf{z}_i$ = $W^l \mathbf{h}_i^{(l)}$, and $W^{(l)}$ is the weight matrix in layer $l$.

This theorem establishes that minimizing representation disparities between demographic groups provides theoretical guarantees for downstream fairness. Building upon this theoretical foundation, we propose a fairness regularizer to mitigate structural and feature disparities simultaneously. Specifically, we propose a masking-based fairness regularizer that acts on both the masked aggregated representations and the masked node features. For each node $v_i$, we learn a masking vector $\mathbf{m}_i$ = $[m_{i,1}, m_{i,2}, \ldots, m_{i,d_h}]$ and form:

\begin{equation}
\label{eq:masked_repr}
\tilde{\mathbf{h}}_i = \mathbf{h}_i \odot \mathbf{m}_i = [h_{i,1} m_{i,1}, h_{i,2} m_{i,2}, \ldots, h_{i,d_h} m_{i,d_h}]
\end{equation}

\noindent where $\odot$ denotes element-wise product, and $\tilde{\mathbf{h}}_i$ represent the fair node representation of node $v_i$.


The masking vectors are learned by minimizing an MMD-based bias measure on masked representations:

\begin{align}
\label{eq:masked_repr}
\mathcal{L}_{F} &= \frac{1}{N_d^2} \sum_{v_i \in \mathcal{S}_d} \sum_{v_j \in \mathcal{S}_d} k(\tilde{\mathbf{h}}_i, \tilde{\mathbf{h}}_j) + \frac{1}{N_f^2} \sum_{v_i \in \mathcal{S}_f} \sum_{v_j \in \mathcal{S}_f} k(\tilde{\mathbf{h}}_i, \tilde{\mathbf{h}}_j) \\ \nonumber
&- \frac{2}{N_d N_f} \sum_{v_i \in \mathcal{S}_d} \sum_{v_j \in \mathcal{S}_f} k(\tilde{\mathbf{h}}_i, \tilde{\mathbf{h}}_j)
\end{align}

\noindent where $k(\cdot)$ denotes the differentiable positive semidefinite kernels (\textit{e.g.}, RBF kernel).

Furthermore, to avoid scenarios where the model collapses nodes into a single subgroup instead of genuinely reducing disparities between different subgroups, an additional penalty term is introduced to explicitly discourage representations highly correlated with demographic labels:

\begin{equation}
    \mathcal{L}_R = \sum_{c=1}^{d_h} \left(\text{Cov}(s, \hat{h}_{s,c})\right)^2
\end{equation}

\noindent where $\text{Cov}(\cdot)$ denotes the covariance.

In summary, the fairness constraint effectively minimizes representation disparities between demographic subgroups and simultaneously penalizes excessive correlations with demographic labels. The integration of both the masking-based regularizer and the covariance penalty ensures that nodes from different demographic groups achieve similar masked representations, significantly reducing the influence of demographic information on downstream predictions while maintaining the capacity to encode task-relevant patterns.

\noindent \textbf{Information Constraint.} For each node $v_i$, the fair node representation $\tilde{\mathbf{h}}_{v_i}$ should preserve essential features and structural information, ensuring its usefulness for downstream tasks. In other words, the model should be able to make accurate label predictions using node representations (\textit{i.e.}, $\tilde{\mathbf{h}}_{v_i} \rightarrow y_{v_i}$). As illustrated in Figure~\ref{fig:overview}, the information constraint ensures the retention of the task-related information to accurately predict the label in both the fair node embedding $\tilde{\mathbf{h}}_{v_i}$ and the original node embedding $\mathbf{h}_{v_i}$. Hence, the objective of the information constraint is to minimize the loss of the prediction model, as shown in Equation~\ref{equ:Cp}:

\begin{equation}
\label{equ:Cp}
\mathcal{L}_I = \frac{1}{|\mathcal{V}_L|} \sum_{v_i \in \mathcal{V}_L} - (y_{v_i} \log(\hat{y_{v_i}}) + (1 - y_{v_i}) \log(1-\hat{y_{v_i}}))
\end{equation}

\noindent where $y_i$ is the one-hot encoding of the ground-truth label of node $v_i$, and $\hat{y}_{v_i}$ denotes the predicted probability of the correct label derived from the fair node representations.

\noindent \textbf{Graph Reconstruction Constraint.} For each node $u$, another objective is to ensure that its node representations accurately represent the node itself. This requirement is fulfilled by accurately reconstructing the ego-graph $\mathcal{G}_{v_i}$ from the new node embedding $\mathrm{\mathbf{h'}}_i$. Formally, we define the graph reconstruction constraint as a graph structure reconstruction loss, $\mathcal{L}_G$:

\begin{equation}
\label{equ:Jr}
\mathcal{L}_G 
=\; 
\frac{1}{|\mathcal{E}_{S_d}| + |\mathcal{E}_{S_f}|} 
\sum_{e_{i,j} \in \mathcal{E}} 
L\bigl(e_{i,j}, \hat{e}_{i,j}\bigr)
\end{equation}

\noindent where $\mathcal{E}_{S_d}$ and $\mathcal{E}_{S_f}$ are sets of sampled edges connecting nodes from deprived and favored subgroups, respectively, and $L(\cdot)$ is the cross-entropy loss. The term $e_{i,j}$ denotes the actual connection status between nodes $u$ and $k$, whereas $\hat{e}_{i,j} = \sigma(\mathrm{\mathbf{h'}}_i {\mathrm{\mathbf{h'}}}_j^\top)$ is the predicted probability of a link, with $\sigma(\cdot)$ representing the sigmoid function. 

In summary, the graph reconstruction constraint helps preserve critical structural information in node embeddings, effectively preventing bias information. This ensures that the reconstructed ego-graph $\mathcal{G}_{v_i}$ remains consistent with the original graph topology, thereby accurately reflecting structural relationships while reducing demographic biases.

\subsection{Adaptivity Confidence Strategy Module}
\label{sec:adaptivity_confidence_strategy_module}

Building on the previous two modules, this module entails an adaptive confidence strategy that intelligently modulates fairness enforcement based on the reliability of demographic predictions. The key insight driving this module is that fairness constraints should be applied more stringently to nodes where we have high confidence in their demographic group membership, while being more lenient with nodes whose demographic information remains uncertain. The rationale for this confidence-based weighting stems from the fundamental challenge of working with incomplete demographic information. When our demographic identification module makes highly confident predictions about a node's demographic information, we can trust these demographic proxies and should therefore enforce strict fairness constraints to prevent bias. However, when the demographic predictions are uncertain, applying strong fairness penalties may be counterproductive, as we might be enforcing constraints based on potentially incorrect demographic assignments.

To implement this adaptive approach, we define a confidence score ($\mathrm{conf}(v_i)$), for each node $v_i$, representing the reliability of the predicted demographic label. A predefined threshold $\tau$ is then utilized to categorize nodes into high-confidence and low-confidence groups. Nodes with confidence scores greater than or equal to $\tau$ are considered reliable, thus warranting stronger fairness enforcement to mitigate embedded biases. Conversely, nodes with confidence scores below are treated as uncertain, and fairness constraints are applied more leniently to avoid potential errors from inaccurate demographic predictions. Formally, the fairness loss can be expressed as:

\begin{equation}
\mathcal{L}_{F}
=
\sum_{i,i' \in \mathcal{S}_d} \alpha_i \alpha_{i'} \, k(\tilde{\mathbf{h}}_i, \tilde{\mathbf{h}}_{i'})
\;+\;
\sum_{j,j' \in \mathcal{S}_f} \beta_j \beta_{j'} \, k(\tilde{\mathbf{h}}_j, \tilde{\mathbf{h}}_{j'})
\;-\;
2 \sum_{i \in \mathcal{S}_d} \sum_{j \in \mathcal{S}_f} \alpha_i \beta_j \, k(\tilde{\mathbf{h}}_i, \tilde{\mathbf{h}}_j),
\end{equation}

\noindent where the confidence weights $\alpha_i=\frac{\mathrm{conf}(v_i)}{\sum_{v_k\in\mathcal{S}_d}\mathrm{conf}(v_k)}$ for $v_i\in\mathcal{S}_d$, and $\beta_j=\frac{\mathrm{conf}(v_j)}{\sum_{v_k\in\mathcal{S}_f}\mathrm{conf}(v_k)}$ for $v_j\in\mathcal{S}_f$ are normalized within each predicted demographic group, ensuring that nodes with high-confidence demographic predictions receive proportionally more emphasis in the fairness calculation, while nodes with uncertain demographic assignments contribute less to the fairness loss.



In addition to the adaptive fairness loss, an adaptive correlation penalty $\mathcal{L}_R$ is introduced to further strengthen fairness by explicitly reducing correlations between node representations and demographic labels. This penalty is formally defined as:

\begin{equation}
    \mathcal{L}_R = \sum_{c=1}^{d_h} \left( \frac{1}{W} \sum_{i=1}^{n} w_i \left( s_i - \frac{1}{W} \sum_{u=1}^{n} w_u s_u \right) \left( \tilde{h}_{i,c} - \frac{1}{W} \sum_{v=1}^{n} w_v \tilde{h}_{v,c} \right) \right)^2
\end{equation}

In this way, nodes with high-confidence predictions receive more stringent fairness treatment, ensuring that their label prediction does not induce bias. Meanwhile, nodes with low-confidence predictions incur a smaller fairness penalty, acknowledging that the classifier's uncertainty already diminishes the likelihood of discrimination arising from their representations while also helping the model increase the confidence of its predictions. In summary, the adaptivity confidence strategy module first focuses on nodes whose demographic information has been reliably identified, accurately enforcing stronger fairness constraints for these high-confidence cases. Meanwhile, for nodes with initially uncertain demographic information, the module imposes milder constraints, allowing the model to progressively refine its predictions. As confidence in demographic identification for these uncertain nodes increases over time, fairness constraints are correspondingly strengthened. Through this dynamic approach, the module systematically enhances model fairness while preserving high prediction accuracy.

\subsection{Overall Learning Object}
\label{sec:learning_object}

To jointly optimize utility and fairness, FairGLite employs a unified loss function within its fair graph representation learning framework under limited demographics. This loss integrates three components: (i) information loss to preserve task-relevant signals, (ii) graph reconstruction loss to recover the graph structure, and (iii) fairness loss to eliminate demographic information. The overall objective is formulated as:

\begin{equation}
\label{eq:total}
\min\ \ \mathcal{L}_{\mathrm{total}}
\;=\;
\mathcal{L}_{I}
\;+\;
a\,\mathcal{L}_{G}
\;+\;
b\,(\mathcal{L}_{F} + \mathcal{L}_{R}),
\end{equation}

\noindent where $a$ and $b$ are tunable hyperparameters to balance the contributions of the various elements in the overall objective function, with $\mathcal{L}_I$, $\mathcal{L}_G$, $\mathcal{L}_F$, and $\mathcal{L}_R$ corresponding to the utility loss, the graph reconstruction loss, the fairness loss, and the correlation penalty loss that explicitly penalizes excessive correlations between node representations and demographic labels, respectively.

\section{Experiment}
\label{sec:experiment}


This section presents the experimental evaluation of FairGLite. We first describe the datasets, baseline methods, and evaluation metrics, followed by the presentation and analysis of results. Due to space constraints, additional evaluations are provided in the appendix\footref{fn:link}.

\subsection{Experimental Settings}

\noindent \textbf{Datasets.} We evaluate FairGLite on four widely used real-world datasets, \textit{i.e.}, the \textbf{Credit} dataset~\cite{yeh2009comparisons}, \textbf{Pokec-z} and \textbf{Pokec-n} datasets~\cite{takac2012data}, and the \textbf{NBA} dataset~\cite{dai2021say}. The \textbf{Credit} dataset consists of credit card holders represented as nodes, connected by edges based on similarities in spending and payment behaviors. Each node includes transaction-related features. The \textbf{Pokec-z} and \textbf{Pokec-n} datasets originate from a popular social network in Slovakia, corresponding to two distinct provincial sub-networks. Nodes represent users characterized by attributes such as gender, age, and interests, while edges represent friendships. The prediction task involves classifying users' occupational fields. The \textbf{NBA} dataset models professional basketball players as nodes, connected based on similarity in performance metrics. The prediction task is to determine if a player's salary exceeds the league average. The detailed statistics of these datasets are shown in Table~\ref{tab:dataset_info}. In all datasets, isolated nodes are removed before experiments. The data is partitioned into training (50\%), validation (20\%), and testing (30\%) sets. To evaluate the effectiveness of our method under scenarios with incomplete demographic information, we randomly select 40\% of nodes in the training and validation sets and mask their demographic labels, while maintaining complete labeling of demographic information in the testing set.

\vspace{-0.5cm}
\begin{table*}
\centering
\caption{Summary of the datasets in the experiments.}
\label{tab:dataset_info}
\begin{adjustbox}{scale=1}
\begin{tabular}{>{\centering\arraybackslash}m{0.8in}>{\centering\arraybackslash}m{0.5in}>{\centering\arraybackslash}m{0.5in}>{\centering\arraybackslash}m{0.5in}>{\centering\arraybackslash}m{0.5in}}
\toprule
\textbf{Dataset}  & \textbf{Credit}& \textbf{Pokec-z} & \textbf{Pokec-n}& \textbf{NBA} \\
\midrule
Vertices   & 30,000 &67,797 &66,569 & 403 \\
\midrule
Edges    & 137,377 &882,765 &729,129 & 16,570 \\
\midrule
Feature Dimension   &13 &65 &65 & 97 \\
\midrule
Demographics  & Age &Region&Region & Country\\
\bottomrule
\end{tabular}  
\end{adjustbox}
\end{table*}
\vspace{-0.3cm}

\noindent \textbf{Baselines.} We compare FairGLite with several state-of-the-art methods, grouped into two categories. \textbf{(i) Vanilla Graph Model:} GCN~\cite{kipf2016semi}, which applies spectral graph convolutions without fairness constraints. \textbf{(ii) Fairness-aware Methods:} FairKD~\cite{chai2022fairness}, which trains a teacher model to overfit and generate soft labels that guide a student model via knowledge distillation; KSMOTE~\cite{yan2020fair}, which uses clustering to assign proxy demographic labels and balances subgroups through synthetic oversampling; FairRF~\cite{zhao2022towards}, which mitigates feature-related biases directly without requiring demographic attributes; FairAC~\cite{guo2023fair}, which extends fairness to graph data by embedding nodes with observed attributes and using attention to aggregate neighbor information for missing attributes; FairGKD~\cite{zhu2024devil}, which transfers fair representations learned by a teacher GNN to a student model via graph-based knowledge distillation; and FairGNN~\cite{dai2021say}, which estimates demographics and improves fairness through adversarial learning under limited demographic information. For methods not originally designed for graphs (FairKD, KSMOTE, FairRF), we adapt their implementations to our backbone using the authors’ released code.

\noindent \textbf{Evaluation Metrics.} We evaluated the proposed framework with respect to two key aspects: prediction performance and fairness performance. To evaluate prediction performance, we chose two metrics for node classification, \textit{i.e.}, accuracy and F1-Score~\cite{utilitymetrics}, where higher scores indicate better prediction results. For fairness assessment, we utilize two commonly used metrics: Demographic parity ($\Delta_{DP}$)~\cite{le2022survey} and Equal Opportunity ($\Delta_{EO}$)~\cite{hardt2016equality}. These fairness metrics measure the disparity in predictions between different demographic groups, where values closer to zero indicate higher fairness.

\vspace{-1cm}

\begin{table*}[!h]
\centering
\caption{Comparison of FairGLite with baselines (columns are metrics; rows are methods). The best in each metric column is \textbf{bold}, second-best is \underline{underlined}.}
\vspace{0.2cm}
\label{tab:RQ1_transposed}
\begingroup

\setlength{\tabcolsep}{8pt}      
\renewcommand{\arraystretch}{1.12}
\begin{adjustbox}{scale=0.92}
\begin{tabular}{|c||c|c|c|c|c|}
\Xhline{1.2pt}
\textbf{Dataset} & \textbf{Method} & \textbf{Accuracy (\(\uparrow\))} & \textbf{F1-Score (\(\uparrow\))} & \(\boldsymbol{\Delta_{DP}}(\downarrow)\) & \(\boldsymbol{\Delta_{EO}}(\downarrow)\) \\
\Xhline{1.2pt}

\multirow{8}{*}{\textbf{Credit}}
& GCN      & \textbf{0.781 \(\pm\) 0.016} & \textbf{0.868 \(\pm\) 0.023} & 0.117 \(\pm\) 0.013 & 0.096 \(\pm\) 0.017 \\
& KSMOTE   & 0.736 \(\pm\) 0.009 & 0.817 \(\pm\) 0.012 & 0.071 \(\pm\) 0.003 & 0.055 \(\pm\) 0.013 \\
& FairKD   & 0.711 \(\pm\) 0.012 & 0.796 \(\pm\) 0.023 & 0.094 \(\pm\) 0.036 & 0.075 \(\pm\) 0.042 \\
& FairRF   & 0.735 \(\pm\) 0.017 & 0.809 \(\pm\) 0.022 & 0.067 \(\pm\) 0.017 & 0.057 \(\pm\) 0.018 \\
& FairAC   & \underline{0.748 \(\pm\) 0.026} & 0.831 \(\pm\) 0.018 & 0.047 \(\pm\) 0.015 & 0.041 \(\pm\) 0.014 \\
& FairGKD  & 0.743 \(\pm\) 0.028 & \underline{0.834 \(\pm\) 0.013} & \underline{0.038 \(\pm\) 0.011} & \underline{0.037 \(\pm\) 0.021} \\
& FairGNN  & 0.687 \(\pm\) 0.012 & 0.783 \(\pm\) 0.043 & 0.123 \(\pm\) 0.026 & 0.115 \(\pm\) 0.02\\
\Xcline{2-6}{0.8pt}
& \textbf{FairGLite}& 0.743 \(\pm\) 0.032 & 0.825 \(\pm\) 0.018 & \textbf{0.035 \(\pm\) 0.015} & \textbf{0.033 \(\pm\) 0.013} \\
\Xhline{1.1pt}

\multirow{8}{*}{\textbf{Pokec-z}}
& GCN      & \textbf{0.699 \(\pm\) 0.024} & \textbf{0.622 \(\pm\) 0.024} & 0.075 \(\pm\) 0.025 & 0.062 \(\pm\) 0.013 \\
& KSMOTE   & 0.697 \(\pm\) 0.024 & 0.611 \(\pm\) 0.018 & 0.037 \(\pm\) 0.017 & 0.039 \(\pm\) 0.010 \\
& FairKD   & 0.673 \(\pm\) 0.021 & 0.592 \(\pm\) 0.013 & 0.045 \(\pm\) 0.014 & 0.048 \(\pm\) 0.009 \\
& FairRF   & 0.690 \(\pm\) 0.014 & 0.617 \(\pm\) 0.019 & 0.032 \(\pm\) 0.012 & 0.034 \(\pm\) 0.012 \\
& FairAC   & 0.655 \(\pm\) 0.031 & 0.603 \(\pm\) 0.013 & 0.032 \(\pm\) 0.018 & \underline{0.029 \(\pm\) 0.014} \\
& FairGKD  & 0.660 \(\pm\) 0.025 & 0.618 \(\pm\) 0.009 & \textbf{0.029 \(\pm\) 0.021} & 0.030 \(\pm\) 0.018 \\
& FairGNN  & 0.689 \(\pm\) 0.071 & 0.603 \(\pm\) 0.021 & 0.038 \(\pm\) 0.022 & 0.033 \(\pm\) 0.029 \\
\Xcline{2-6}{0.8pt}
& \textbf{FairGLite}& \underline{0.671 \(\pm\) 0.041} & \underline{0.620 \(\pm\) 0.032} & \underline{0.031 \(\pm\) 0.013} & \textbf{0.027 \(\pm\) 0.015} \\
\Xhline{1.1pt}

\multirow{8}{*}{\textbf{Pokec-n}}
& GCN      & \underline{0.689 \(\pm\) 0.015} & \textbf{0.631 \(\pm\) 0.022} & 0.084 \(\pm\) 0.013 & 0.078 \(\pm\) 0.019 \\
& KSMOTE   & 0.669 \(\pm\) 0.013 & 0.611 \(\pm\) 0.018 & 0.061 \(\pm\) 0.010 & 0.066 \(\pm\) 0.013 \\
& FairKD   & 0.663 \(\pm\) 0.016 & 0.603 \(\pm\) 0.023 & 0.067 \(\pm\) 0.015 & 0.064 \(\pm\) 0.013 \\
& FairRF   & 0.673 \(\pm\) 0.013 & 0.616 \(\pm\) 0.032 & 0.056 \(\pm\) 0.009 & 0.061 \(\pm\) 0.016 \\
& FairAC   & 0.675 \(\pm\) 0.028 & 0.621 \(\pm\) 0.026 & 0.026 \(\pm\) 0.013 & \underline{0.030 \(\pm\) 0.027} \\
& FairGKD  & 0.681 \(\pm\) 0.021 & 0.628 \(\pm\) 0.029 & \underline{0.025 \(\pm\) 0.015} & 0.035 \(\pm\) 0.030 \\
& FairGNN  & 0.675 \(\pm\) 0.028 & 0.619 \(\pm\) 0.032 & 0.036 \(\pm\) 0.012 & 0.044 \(\pm\) 0.020 \\
\Xcline{2-6}{0.8pt}
& \textbf{FairGLite}& \textbf{0.689 \(\pm\) 0.024} & \underline{0.630 \(\pm\) 0.029} & \textbf{0.023 \(\pm\) 0.018} & \textbf{0.028 \(\pm\) 0.013} \\
\Xhline{1.1pt}

\multirow{8}{*}{\textbf{NBA}}
& GCN      & 0.668 \(\pm\) 0.025 & 0.703 \(\pm\) 0.022 & 0.063 \(\pm\) 0.043 & 0.074 \(\pm\) 0.043 \\
& KSMOTE   & 0.654 \(\pm\) 0.023 & 0.685 \(\pm\) 0.038 & 0.057 \(\pm\) 0.033 & 0.065 \(\pm\) 0.033 \\
& FairKD   & 0.671 \(\pm\) 0.036 & 0.681 \(\pm\) 0.023 & 0.042 \(\pm\) 0.025 & 0.055 \(\pm\) 0.014 \\
& FairRF   & 0.664 \(\pm\) 0.033 & 0.687 \(\pm\) 0.012 & 0.044 \(\pm\) 0.038 & 0.042 \(\pm\) 0.026 \\
& FairAC   & \textbf{0.673 \(\pm\) 0.028} & 0.699 \(\pm\) 0.038 & \underline{0.035 \(\pm\) 0.009} & 0.037 \(\pm\) 0.017 \\
& FairGKD  & 0.670 \(\pm\) 0.024 & \underline{0.706 \(\pm\) 0.033} & 0.040 \(\pm\) 0.067 & \underline{0.032 \(\pm\) 0.010} \\
& FairGNN  & 0.658 \(\pm\) 0.027 & 0.694 \(\pm\) 0.032 & 0.036 \(\pm\) 0.021 & 0.034 \(\pm\) 0.025 \\
\Xcline{2-6}{0.8pt}
& \textbf{FairGLite}& \underline{0.723 \(\pm\) 0.024} & \textbf{0.711 \(\pm\) 0.029} & \textbf{0.032 \(\pm\) 0.036} & \textbf{0.030 \(\pm\) 0.005} \\
\Xhline{1.2pt}
\end{tabular}
\end{adjustbox}
\endgroup
\label{tab:comparison}
\end{table*}

\subsection{Experimental Results}

\noindent \textbf{RQ1: How does FairGLite perform in balancing utility and fairness across real-world graph datasets?} To answer this question, Table~\ref{tab:comparison} summarizes the comparisons between our proposed method, FairGLite, and the baseline methods. Specifically, two key observations emerge: i) FairGLite achieves superior fairness when demographic information is missing. Across all evaluated datasets, FairGLite consistently shows better fairness performance than baseline methods. This advantage stems from FairGLite's ability to effectively leverage node features and graph structure to accurately generate proxy of demographic information, establishing a solid foundation for bias mitigation. Furthermore, FairGLite mitigates multiple forms of bias in graph data, better preventing demographic information from leaking into downstream classification tasks. ii) FairGLite demonstrates comparable predictive performance compared with existing fairness methods. Unlike existing approaches that impose uniform fairness constraints on all nodes, FairGLite dynamically adjusts each node's contribution to the fairness loss through the adaptivity confidence strategy, enabling better learning for nodes with low confidence. Overall, these results highlight FairGLite's advantage in effectively balancing predictive performance and fairness.

\begin{figure}[h]
	\centering
	\includegraphics[width=0.95\textwidth]{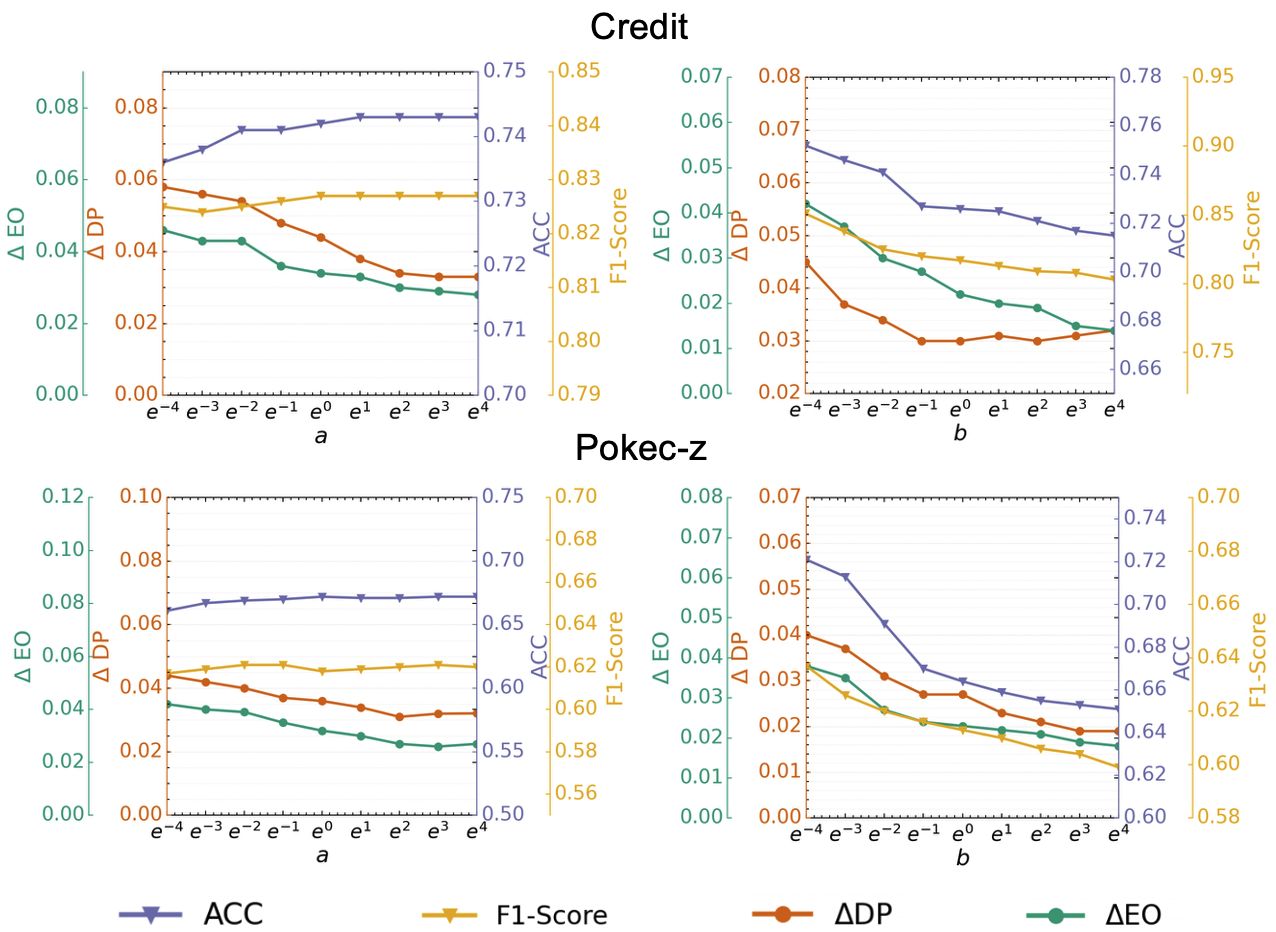}
	\caption{Study on Hyper-parameters sensitivity analysis.}
	\label{fig:parameters}
\end{figure}

\noindent \textbf{RQ2: How Do the Hyper-parameters $a$ and $b$ Impact the Trade-off Between Utility and Fairness in FairGLite?} We investigate the sensitivity of FairGLite to two key hyperparameters, \textit{i.e.}, $a$ and $b$. As shown in Figure~\ref{fig:parameters}, as $a$ increases, the model achieves better prediction performance and fairness. However, if it passes a certain threshold, both prediction performance and fairness stabilize. For parameter $b$, as shown in Figure~\ref{fig:parameters}, we observe three distinct phases: when $b$ is very small, the fairness constraints have minimal impact. As $b$ increases, fairness steadily improves, though prediction accuracy gradually declines due to stronger regularization. Beyond a threshold (\textit{e.g.}, $e^{1}$ for Credit/NBA, $e^{3}$ for Pokec-z/Pokec-n), fairness performance stabilizes or slightly deteriorates because excessive regularization restricts the model's representational capacity. Note that results for NBA and Pokec-n are provided in the appendix\footref{fn:link}. To sum up, these results highlight the trade-off between fairness and task performance. Thus, careful tuning of $a$ and $b$ is essential for optimal model performance.

\noindent \textbf{RQ3: What is the Impact of Each Component on the FairGLite on its utility and fairness?} We conducted ablation studies to assess the contributions of each module within the FairGLite framework. FairGLite consists of three key modules: the Demographic Information Identification Module, the Fair Node Representation Learning Module, and the Adaptivity Confidence Strategy Module. Notably, we did not create a variant without the Demographic Information Identification Module because this module is foundational to FairGLite's operation. Without it, the framework cannot identify proxies for the missing demographic information, which are essential inputs for the subsequent modules. Therefore, removing this component would render the entire framework inoperable, making such an ablation study impractical. We present the ablation study results for the Credit and Pokec-z datasets in Figure~\ref{fig:ablation}, additional results for NBA and Pokec-n datasets are provided in the appendix\footref{fn:link}.

\begin{figure}[h]
	\centering
 	\includegraphics[width=1\textwidth]{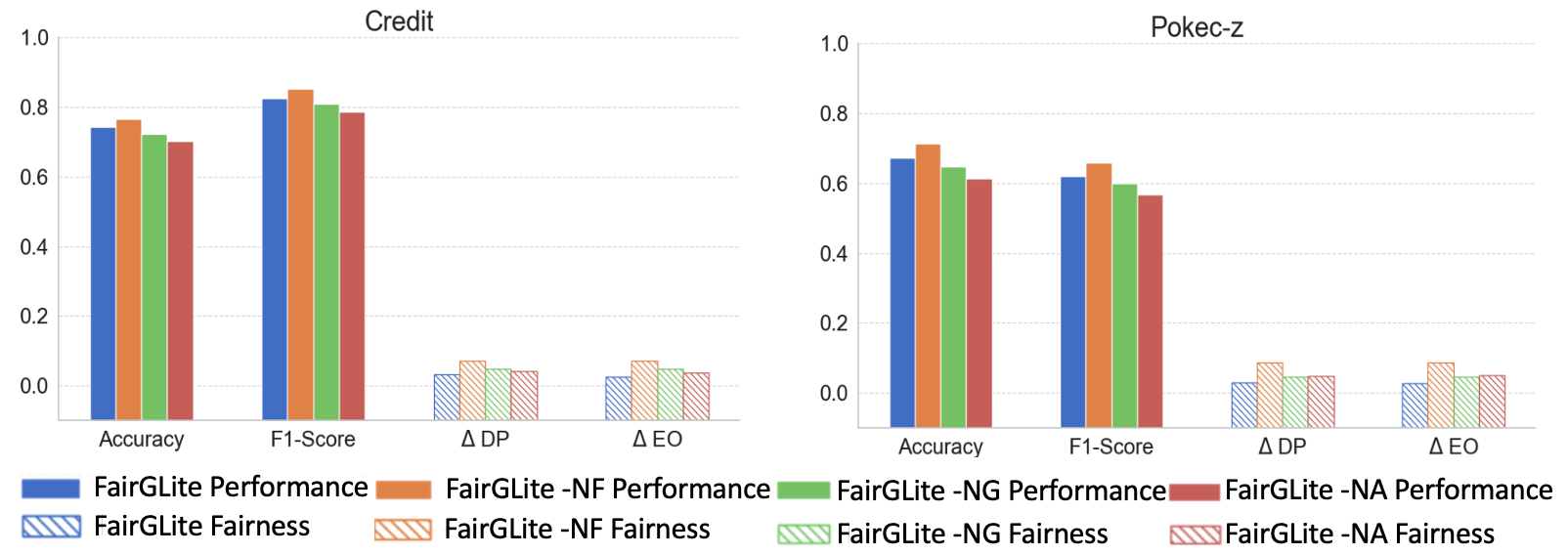}
	\caption{Ablation study results for FairGLite, FairGLite-NF, FairGLite-NG and FairGLite-NA.}
	\label{fig:ablation}    
\end{figure}

For the Fair Node Representation Learning Module, we created two variants: FairGLite-NF (without the Fairness Constraint) and FairGLite-NG (without the Graph Reconstruction Constraint). As shown in Figure~\ref{fig:ablation}, the fairness metrics of FairGLite-NF dropped significantly. This occurs because without the Fairness Constraint, demographic information in node representations directly passes to downstream classification tasks, leading to discriminatory decisions. The FairGLite-NG variant shows better fairness metrics than FairGLite-NF but still demonstrates a slight decrease compared to the full FairGLite model, along with reduced performance metrics. This degradation occurs because without the Graph Reconstruction Constraint, node representations fail to capture important structural information, resulting in decreased graph representational performance.

We also examined the impact of the Adaptivity Confidence Strategy Module by creating the FairGLite-NA variant (without adaptive confidence). As shown in Figure~\ref{fig:ablation}, FairGLite-NA shows reduced performance compared to the complete FairGLite model. This is because applying fairness constraints with equal strength to all nodes makes it more difficult for the model to learn from samples with low confidence, thereby reducing the overall accuracy.

\section{Conclusion}
\label{sec:conclusion}

This paper addresses the critical gap between real-world constraints on demographic information availability and the assumptions underlying existing fairness-aware graph learning methods. To bridge this gap, we proposed FairGLite, a theoretically grounded framework designed to promote fairness in graph-based decision-making. FairGLite effectively mitigates bias in node representations without depending on complete demographic information. In addition, we proposed an adaptive confidence strategy that further enhances its practical utility, intelligently balancing fairness and predictive accuracy. Through rigorous theoretical analysis, we demonstrated the robustness of FairGLite, establishing its potential as a versatile and broadly applicable solution for real-world fair graph learning.

\section*{Acknowledgements}

This work was supported in part by the National Science Foundation (NSF) under Grant No. 2404039 and the National Institutes of Health (NIH) under Grant No. R01MD019814.

\end{document}